\documentclass{article}

\usepackage{PRIMEarxiv}

\usepackage[utf8]{inputenc} 
\usepackage[T1]{fontenc}    
\usepackage{hyperref}       
\usepackage{url}            
\usepackage{booktabs}       
\usepackage{amsfonts}       
\usepackage{nicefrac}       
\usepackage{microtype}      
\usepackage{lipsum}
\usepackage{fancyhdr}       
\usepackage{graphicx}       
\graphicspath{{media/}}     

\usepackage{graphicx}
\usepackage{subcaption}
\usepackage{amsmath}
\usepackage{amssymb}
\usepackage{tabularx}
\usepackage{makecell}
\usepackage[most]{tcolorbox}
\usepackage{enumitem}
\usepackage{multirow}
\usepackage{CJKutf8}

\usepackage{algorithm}
\usepackage{algpseudocode} 

\usepackage{amsmath}
\usepackage{amssymb}
\usepackage{mathtools}
\usepackage{amsthm}
\usepackage{float}

\usepackage[most]{tcolorbox}
\usepackage{xcolor}

\title{CirrusBench: Evaluating LLM-based Agents Beyond Correctness in Real-World Cloud Service Environments
}

\author{
  Yi Yu$^{1*}$, Guangquan Hu$^{4*}$, Chenghuang Shen$^{2*,3}$,Xingyan Liu$^{3*}$,\\
  Jing Gu$^{1\dagger}$, Hangyi Sun$^{1\dagger}$, Junzhuo Ma$^{1}$,Weiting Liu$^{4}$ , \\
  Jianfeng Liu$^{3\ddagger}$, Mingyue Pu$^{3}$, Yu Wang$^{3}$, \\
  Zhengdong Xiao$^{3}$, Rui Xie$^{3}$, Longjiu Luo$^{3}$, Qianrong Wang$^{3}$, \\
  Gurong Cui$^{3}$, Honglin Qiao$^{3}$, Wenlian Lu$^{1,2,4,5\ddagger}$ \\
  \\
  $^{1}$School of Mathematics and Sciences, Fudan University, Shanghai, China \\
  $^{2}$Shanghai Center for Mathematical Sciences, Fudan University, Shanghai, China \\
  $^{3}$Alibaba Group, Hangzhou, China \\
  $^{4}$Institute of Science and Technology for Brain-Inspired Intelligence, Fudan University, Shanghai, China \\
  $^{5}$Center for Applied Mathematics \& Shanghai Key Laboratory of Contemporary Applied Mathematics, \\
  Fudan University, Shanghai, China \\
  \texttt{jiawei.ljf@alibaba-inc.com} \\
  \texttt{wenlian@fudan.edu.cn} \\
  \\
  {\small $^{*}$Contributed equally to this research.  $^{\dagger}$Contributed equally to this research.$^{\ddagger}$Corresponding author.}
}

\begin{document}
\maketitle

\begin{abstract}
   The increasing agentic capabilities of Large Language Models (LLMs) have enabled their deployment in real-world applications, such as cloud services, where customer-assistant interactions exhibit high technical complexity and long-horizon dependencies, making robustness and resolution efficiency critical for customer satisfaction. However, existing benchmarks for LLM-based agents largely rely on synthetic environments that fail to capture the diversity and unpredictability of authentic customer inputs, often ignoring the resolution efficiency essential for real-world deployment. To bridge this gap, we introduce \textbf{CirrusBench}\footnote{CirrusBench evaluation framework is released at: \url{https://github.com/CirrusAI}}, a novel evaluation framework distinguished by its foundation in real-world data from authentic cloud service tickets. \textbf{CirrusBench} preserves the intricate multi-turn logical chains and realistic tool dependencies inherent to technical service environments. Moving beyond execution correctness, we introduce novel Customer-Centric metrics to define agent success, quantifying service quality through metrics such as the \textit{Normalized Efficiency Index} and \textit{Multi-Turn Latency} to explicitly measure resolution efficiency. Experiments utilizing our framework reveal that while state-of-the-art models demonstrate strong reasoning capabilities, they frequently struggle in complex, realistic multi-turn tasks and fail to meet the high-efficiency standards required for customer service, highlighting critical directions for the future development of LLM-based agents in practical technical service applications.
\end{abstract}


\section{Introduction}
\label{sec:intro}

The rapid advancement of Large Language Models (LLMs) has enabled their deployment as autonomous agents in complex, real-world applications \cite{team2025kimi, zeng2025glm, team2025longcat}. For instance, Cloud service represents a typical domain of significant difficulty and commercial importance. Interactions in this setting are characterized by high technical complexity, long-horizon dependencies, and a need for sophisticated tool use. Consequently, the agent's robustness and efficiency are critical determinants of customer satisfaction \cite{bermbach2017cloud}.

Recent progress in evaluating LLM agents has been substantial, with benchmarks establishing standards for execution correctness and tool-selection capabilities \cite{budzianowski2018multiwoz, liu2023agentbench, qin2023toolllm}. However, current evaluation methodologies suffer from two critical limitations that impede progress toward real-world viability. First, they largely rely on synthetic environments or role-playing simulations \cite{farn2023tooltalk, qian2025userbench, yao2024tau}. While controllable, these settings fail to capture the diversity, unpredictability, and "noise" of authentic customer-assistant interactions, leading to an overestimation of agent performance in practical scenarios. Second, existing evaluations prioritize task completion, often ignoring the resolution efficiency paramount to service quality. In Cloud service, a correct solution delivered through a prolonged and tedious interaction is a failure from the customer's perspective \cite{bermbach2017cloud}.

To bridge this gap between academic benchmarks and real-world requirements, we introduce \textbf{CirrusBench}, a novel evaluation framework grounded in the reality of Cloud service technical support. CirrusBench is constructed from a large corpus of authentic cloud service tickets, preserving the intricate multi-turn dialogues, complex tool schemas, and genuine customer inputs inherent to this domain. This foundation allows for a more realistic assessment of agent capabilities.

CirrusBench introduces an evaluation framework that assesses both single-turn subtask evaluation and dynamic multi-turn task evaluation. Crucially, moving beyond the metrics such as success rate (SR) and \textit{Task Progression Rate (TPR)} quantifying correctness, we introduce novel Customer-Centric Metrics, including \textit{Logical Jump (LJ)}, \textit{Normalized Efficiency Index (NEI)}, \textit{Single-Turn Latency (STL)} and \textit{Multi-Turn Latency (MTL)}, to quantify the customer satisfactory. These metrics explicitly measure the responsiveness and efficiency that define a successful customer interaction. Our main contributions are:
\begin{enumerate}
\item A new benchmark, \textbf{CirrusBench}, built from real-world cloud service data, featuring realistic multi-turn interactions and complex tool dependencies.
\item A suite of \textbf{Customer-Centric Metrics} (LJ, NEI, STL, MTL) designed to measure the operational efficiency and service quality of LLM agents, supplementing traditional correctness-based evaluations.
\end{enumerate}

Experimental analysis reveals a significant gap between SOTA LLMs and real-world Cloud service demands, particularly due to bottlenecks in tool integration. "Thinking" models often incur high latency without proportional improvements in problem solving, while long-horizon interactions suffer from error accumulation and logical fragility. And the knowledge embedding shows importance for agent in Cloud service tasks. Additionally, agents exhibit a tendency toward premature closure, consistently failing to ask necessary clarifying questions during inquiry tasks. Consequently, future development must prioritize resolution efficiency and multi-turn robustness over isolated accuracy metrics.

\section{Related Work}
\label{sec:relatedwork}

Our work is situated at the intersection of LLM agent benchmarking, tool-use evaluation, and customer service automation.

\textbf{Agent Benchmarks and Realism.} The evaluation of LLM agents has evolved from interactions in static web environments like Webshop \cite{yao2022webshop} and WebArena \cite{zhou2023webarena} to more complex domains like software engineering in SWE-bench \cite{jimenez2023swe}. To assess general capabilities, Agentbench \cite{liu2023agentbench} proposed evaluating agents across multiple distinct environments. A persistent challenge, however, is benchmark realism. Most multi-turn benchmarks rely on synthetic data generation via templates \cite{weston2015towards, byrne2019taskmaster} or LLM role-playing \cite{farn2023tooltalk, qian2025userbench, he2025vitabench, zhu2025evaluating}. While scalable, these methods often lack the noisy, diverse, and unpredictable nature of genuine human interactions. A notable exception is the ABCD dataset \cite{chen2021action}, which uses expert live chats. However, it does not incorporate the complex tool-use essential for technical support. CirrusBench addresses this gap by providing a benchmark derived from real-world technical tickets that inherently couples authentic multi-turn dialogue with complex tool integration.

\textbf{Tool Utilization in Multi-turn Interactions.} A significant body of work focuses on evaluating an agent's ability to use tools. Benchmarks like ToolBench \cite{qin2023toolllm}, Metatool \cite{huang2023metatool}, and the 
$\tau$-bench series \cite{yao2024tau, barres2025tau} have systematically assessed tool selection and parameterization. Others, such as ToolSandbox \cite{lu2025toolsandbox}, introduce stateful environments where tool calls dynamically alter the world state. Recent efforts have aimed to embed tool use within conversational contexts \cite{farn2023tooltalk, wangmtu, he2025vitabench}. While these works advance the evaluation of tool-augmented conversation, they are typically built on simulated interactions. In contrast, CirrusBench evaluates an agent's ability to navigate realistic tool dependencies as they organically arise in authentic, long-horizon customer support dialogues.

\textbf{Evaluation Metrics and Customer Satisfaction.} Early evaluation methodologies centered on execution accuracy and keyword matching \cite{zhuang2023toolqa, guo2024ctooleval}. Recognizing the limitations of simple accuracy, researchers have explored LLM-based judges for more nuanced evaluation \cite{mendonccasimple, elizabeth2025neural}. In the specific context of customer service, recent work has begun to formalize metrics for customer satisfaction \cite{lin2024interpretable, hou2025multi}. Moreover, foundational work in cloud service benchmarking has long emphasized that efficiency and latency are critical performance indicators \cite{bermbach2017cloud}. CirrusBench builds upon these ideas by operationalizing customer satisfaction through concrete, objective metrics, such as NEI and MTL, thereby creating the first framework to directly measure the resolution efficiency of LLM agents in a realistic technical support setting.

\section{CirrusBench}
\label{sec:methodology}

To create a benchmark that not only assesses the correctness of an LLM agent but also its efficiency and alignment with real-world customer service workflows, we develop a novel Multi-Turn Task Evaluation framework. Our approach is grounded in the analysis of historical Cloud service tickets and comprises two core components: the metrics measuring correctness and metrics measuring resolution efficiency, reflecting customer satisfactory.

\subsection{Data Definition}

CirrusBench consists of 1,500 rigorously curated tasks without tool-call requirement and 425 tasks with tool-call requirement, systematically derived and refined from authentic dialogue logs obtained from real-world Cloud service operations. These tasks span 20 distinct service categories, including SMS-based support, email correspondence, online technical assistance, and other prototypical service scenarios, thereby ensuring broad coverage of practical inquiry patterns and operational contexts. This diversity of domains enhances the representativeness and ecological validity, providing a robust empirical basis for assessing model performance across heterogeneous service environments. 

\subsection{Data Construction}

\begin{figure}[h!] %
      \centering
      \includegraphics[width=0.9\linewidth]{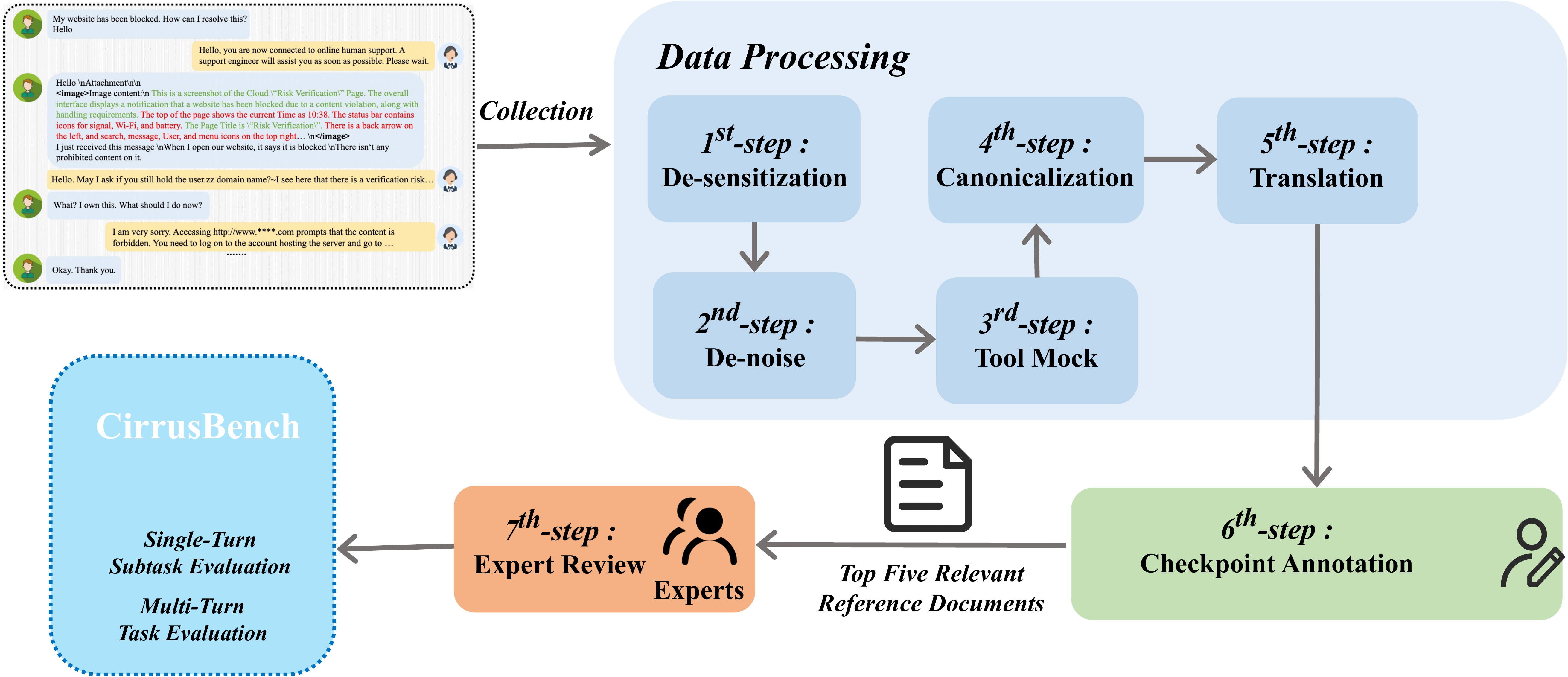}
      \caption{\textbf{The construction of CirrusBench.}    
    The process comprises four core phases: 
    \textit{Phase 1: Data Collection};
    \textit{Phase 2: Preprocessing} (Steps 1--5); 
    \textit{Phase 3: Knowledge-Grounded Annotation} (Step 6); 
    \textit{Phase 4: Quality Assurance} (Step 7). 
    The final benchmark supports both single-turn subtask evaluation and multi-turn task evaluation.}
      \label{fig:construction_Cirrus}
\end{figure}

We began by filtering a large volume of real historical tickets, applying multiple layers of screening through the following steps, and finally converting them into the CirrusBench task-set format. The overall process of construction is shown in Fig. \ref{fig:construction_Cirrus}.

\textbf{Data Collection}

The dataset employed in CirrusBench was systematically constructed from authentic conversational tickets between customers and service assistant, obtained from Cloud service environments. In these live service contexts, customer–assistant interactions typically conform to a canonical problem-resolution schema: customer formulate domain-specific inquiries, and assistant respond with targeted, solution-oriented guidance. 

The tasks were rigorously derived from tickets with a successful resolution. Each task instance encapsulates an entire interaction cycle, extending from the initial customer query to the final confirmation of the solution, such that every ticket corresponds to a genuine, closed-loop service case with explicitly identifiable problem-resolution outcomes. 

This extraction methodology preserves both the structural integrity and contextual authenticity of real-world service exchanges while simultaneously yielding well-defined, self-contained evaluation units that are suitable for systematic and reproducible model assessment.

\subsubsection{Preprocessing}

\textbf{De-sensitization}: Given that CirrusBench is derived from authentic tickets, adhering to strict privacy standards is paramount. Prior to any functional preprocessing, we implemented a rigorous \textit{Personally Identifiable Information (PII) Scrubbing} pipeline. 
We employed a Named Entity Recognition (NER) function to automatically detect and redact sensitive entities, including names, phone numbers, email addresses, and account credentials. All identified PII was replaced with standardized placeholders (e.g., Ja***, 157****3190) to maintain the semantic structure of the dialogue while ensuring full anonymity and compliance with data protection regulations.

\textbf{Denoising and Utterance Pruning}: The raw tickets often contain redundant segments that do not contribute to problem-solving. We performed noise reduction by stripping non-essential conversational fillers, such as repetitive greetings and generic closing formalities. This ensures the model focuses on the core task logic and information exchange.

\textbf{Tool Mock}: A critical challenge in using historical data for tool-use evaluation is the accessibility of live external environments. We address this challenge by adopting a Record-and-Replay methodology to build Tool Mock instances. Specifically, we extracted real-time tool invocations and their corresponding execution results, transforming them into standardized Tool Mock instances. By mapping these to the OpenAI function-calling schema, we enable the benchmark to supply the model with authentic tool outputs without requiring a live connection to backend production APIs, thus ensuring task reproducibility.

\textbf{Dialogue Canonicalization}: In real-world service interfaces, customers frequently transmit their intents across multiple, fragmented messages. To maintain a rigorous Turn-based Interaction framework, we merged consecutive customer queries into single, coherent turns. This process enforces a strict "Query–Reply" (Customer-Assistant) paradigm, ensuring that each assistant response is conditioned on a complete and unambiguous user state.

\textbf{Translation}: We translated the original Chinese tickets into English with an expert translation team.

\textbf{Checkpoint Annotation}: Given the volume of the corpus, we employed a high-capability LLM as an auxiliary annotator to scan the filtered dialogues. The model was prompted to identify turns that involved functional decision-making or information state updates. Specifically, it flagged segments where the assistant’s response was directly tied to the success of the service closed-loop. Beyond knowledge accuracy, we further identify critical \textit{checkpoints} to evaluate the agent's reasoning trajectory. We define checkpoints as critical junctures where the assistant must perform complex reasoning, such as identifying customer intent, deciding to invoke a specific tool, or synthesizing tool outputs into a final solution. To further characterize these interactions, we categorized the response types within these checkpoints into four distinct classes: Solution, Standard, Inquire, and Clarify.

\textbf{Knowledge Augmentation}: For each task in the benchmark, we manually curated and associated at most five highly relevant reference documents. These documents are sourced from official business wikis, standard operating procedures (SOPs), and technical manuals. By providing these documents, we transform the task from a simple text generation problem into a Knowledge-Grounded Generation challenge, requiring the model to synthesize information from external evidence before formulating a response.

\textbf{Quality Control}: We employed several experts in Cloud service to review the constructed dataset, ensuring the quality of CirrusBench.

\subsection{Data Statistics}


\subsubsection{Real-world Service Category distribution}
To ensure the practical utility and robustness of our benchmark, the foundational 1,500 tasks are sourced directly from real-world business operations. Rather than using synthetic conversations, we performed a representative sampling from the top 20 service categrories with the highest ticket frequency in the Cloud service environment. This ensures that the dataset reflects the genuine complexity and linguistic diversity of professional technical support in real-world Cloud service.
The distribution across major service categories is detailed in Table \ref{tab:task_distribution_product}.    

\begin{table}[htbp]

\caption{\textbf{Distribution of Tasks Across Service Categories.} The dataset ($N = 1{,}500$) is sampled from the top 20 service categories with highest ticket frequency in real-world service operations.}
\label{tab:task_distribution_product}
\centering
\begin{tabular}{lcc}
\toprule
\textbf{Category} & \textbf{Task Count} & \textbf{Frequency (\%)} \\ 
\midrule
Server       & 280  & 18.7\% \\
Registration & 220  & 14.7\% \\
SMS Services               & 200  & 13.3\% \\
Website         & 120  & 8.0\%  \\
Email              & 80   & 5.3\%  \\
Others (15+ )    & 600  & 40.0\% \\ 
\midrule
\textbf{Total}             & \textbf{1,500} & \textbf{100.0\%} \\
\bottomrule
\end{tabular}

\end{table}

\begin{figure}[t] %
    \begin{minipage}[t]{0.48\textwidth}
        \includegraphics[width=\linewidth]{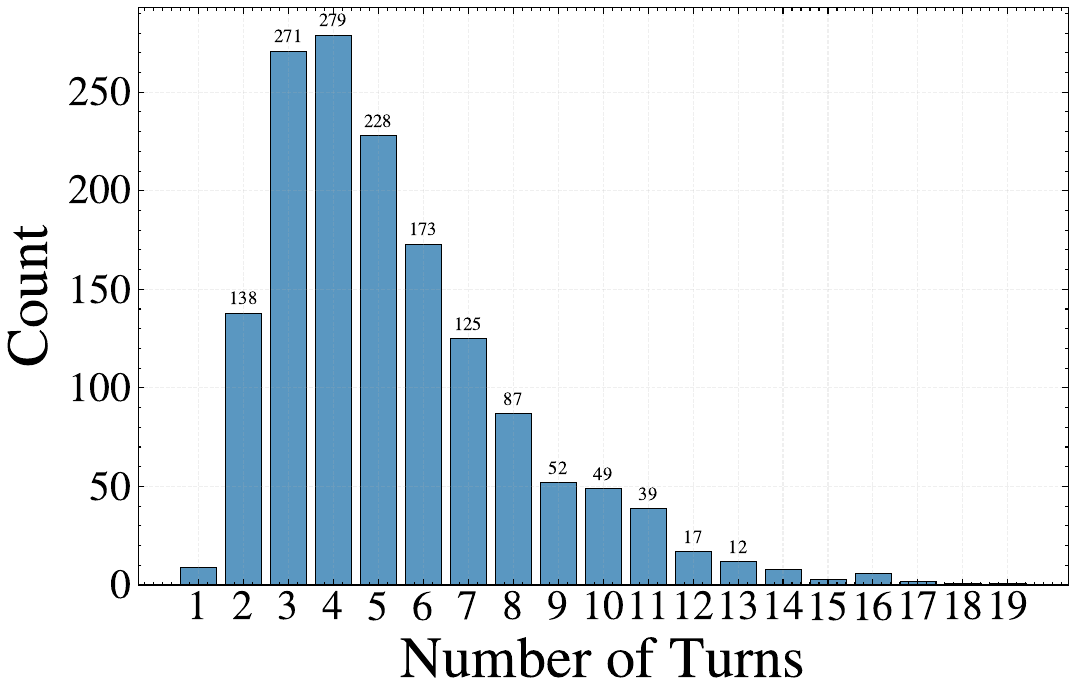}
        \caption{Distribution of the count of conversational turns in preprocessed tickets}
        \label{fig:data_dist_turns}
    \end{minipage}\quad
    \begin{minipage}[t]{0.48\textwidth}
        \includegraphics[width=\linewidth]{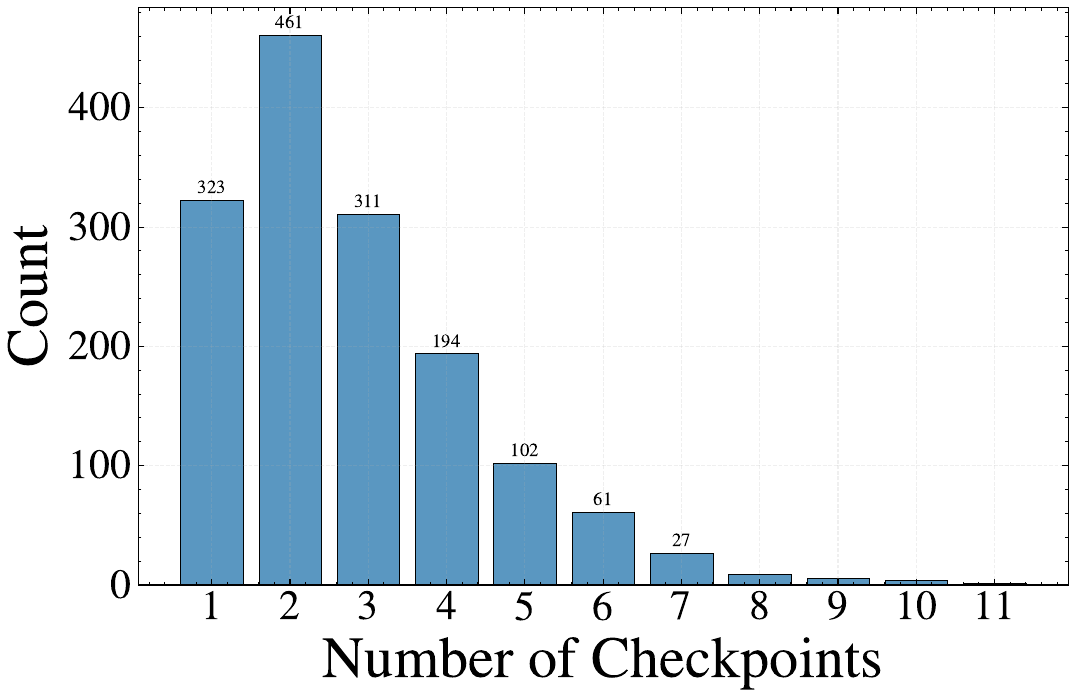}
        \caption{Distribution of the count of checkpoints in processed tickets.}
        \label{fig:data_dist_ckpt}
    \end{minipage}
\end{figure}

\subsubsection{Distribution of interaction complexity}
To characterize the interactive nature and structural depth of the dataset, we analyze the distribution of conversation turns and annotated checkpoints, as illustrated in Figure \ref{fig:data_dist_turns} and Figure \ref{fig:data_dist_ckpt}.

Figure \ref{fig:data_dist_turns} presents the distribution of conversation turns between customer and assistant. The data reveals a significant prevalence of multi-turn interactions, which is a hallmark of complex technical support. 

Figure \ref{fig:data_dist_ckpt} displays the distribution of checkpoints within the conversations. These checkpoints, identified via LLM-based annotation, represent critical reasoning nodes, sub-task transitions, or state verifications necessary to reach a resolution. The variation in the number of checkpoints per instance underscores the logical granularity of the dataset. A higher number of checkpoints indicates a more intricate problem-solving process that demands multi-step reasoning.

The complexity and authenticity of real customer-assistant interactions are also illustrated by examples in the \textbf{Case Study} of \ref{sec:appendix_case}.

These multi-turn dialogues require the model to maintain long-term coherence, track evolving customer requirements, and manage state transitions over extended contexts.

\subsubsection{Distribution of Input Token}
The distribution of input tokens characterizes the dataset's complexity, particularly regarding its suitability for long-context processing. The input sequences exhibit a broad range, extending from a minimum of 520 to a maximum of 37,054 tokens, with a mean length of 11,460.6 tokens.

The distribution reveals significant challenges for contemporary large language models (LLMs). Specifically, the first, second (median), and third quartiles are 6,328.0, 9,777.0, and 14,856.5 tokens, respectively. Notably, with the third quartile (Q3) reaching approximately 15 K tokens and the peak length exceeding 37 K tokens, a substantial portion of the dataset operates in a long-context regime that demands high memory efficiency and robust long-range dependency modeling. The significant gap between the median and the maximum value underscores a "long-tail" distribution of exceptionally dense instances.Moreover, this heterogeneity reflects the inherent diversity and authenticity of customer queries in real-world service environments, where inputs vary from concise requests to highly complex problems, thereby elevating the difficulty of the benchmark beyond simple context length.  These data points provide a rigorous benchmark for evaluating a model's ability to maintain coherence and retrieve information across extensive context windows, ensuring the dataset is well-suited for advancing long-context research.

\subsubsection{Reply types' distribution}
\begin{table}[ht]
\centering
\caption{Distribution of reply types of sub-tasks.}
\label{tab:reply_types}
\begin{tabular}{lrr}
\toprule
\textbf{Reply Type} & \textbf{Count} & \textbf{Proportion (\%)} \\ 
\midrule
Solution            & 2,095          & 63.35\%                 \\
Standard            & 529            & 16.00\%                 \\
Inquire             & 483            & 14.61\%                 \\
Clarify             & 130            & 3.93\%                  \\
Appease             & 70             & 2.12\%                  \\ 
\midrule
\textbf{Total}      & \textbf{3,307} & \textbf{100.00\%}       \\ 
\bottomrule
\end{tabular}
\end{table}

To provide a granular understanding of the functional intent behind the responses, we categorized the sub-task replies annotated by LLM automatedly. The distribution of these reply types is summarized in Table \ref{tab:reply_types}.

The statistics reveal that \textit{solution} is the most prevalent reply type, accounting for 63.35\% (2,095 instances) of the dataset. This dominance aligns with the primary objective of technical support, which is to provide actionable resolutions to customer queries. Beyond direct solutions, the dataset maintains a balanced variety of auxiliary communicative acts: \textit{standard} procedures and \textit{inquire} (seeking more information) constitute 16.00\% and 14.61\% of the replies, respectively. More nuanced interactions, such as \textit{clarify} (3.93\%) and \textit{appease} (2.12\%), are also represented. This diverse distribution ensures that the dataset covers the full spectrum of a typical support dialogue, from initial information gathering and emotional management to final problem resolution, thereby challenging the model to master various conversational strategies in a long-context environment.

\subsubsection{Prevalence and Noise in Image-Derived Content}
The inclusion of screenshots is a defining characteristic of authentic customer-assistant interactions in Cloud services, reflecting the diversity and unpredictability of real-world inputs. In our dataset, 771 out of 1,500 instances (approximately 51.4\%) contain at least one image. To strictly evaluate the reasoning capabilities of text-based LLMs, all visual content in the dataset is transformed into textual format via Optical Character Recognition (OCR), thereby processing the interaction as a uni-modal textual sequence rather than a multimodal one. An example shown in Fig. \ref{fig:case_with_image} can be found in Appendix. \ref{sec:appendix_case}.

However, this transformation preserves the high-entropy nature of the original inputs. Unlike curated synthetic data, real-world screenshots contain significant stochastic noise; they frequently capture extraneous UI elements—such as system timestamps, battery status, navigation bars, and irrelevant background logs—that are semantically orthogonal to the reported technical issue. Since the OCR process transcribes these visual artifacts indiscriminately, the resulting input contains substantial unstructured "textual noise." This shifts the challenge from visual perception to robust information extraction: the model must possess the discriminative capacity to sift through dense, noisy OCR outputs and isolate the specific error context from the surrounding irrelevant information.

\subsection{Evaluation Methodology}

\begin{figure*}[h!] %
      \centering
      \includegraphics[width=0.9\textwidth]{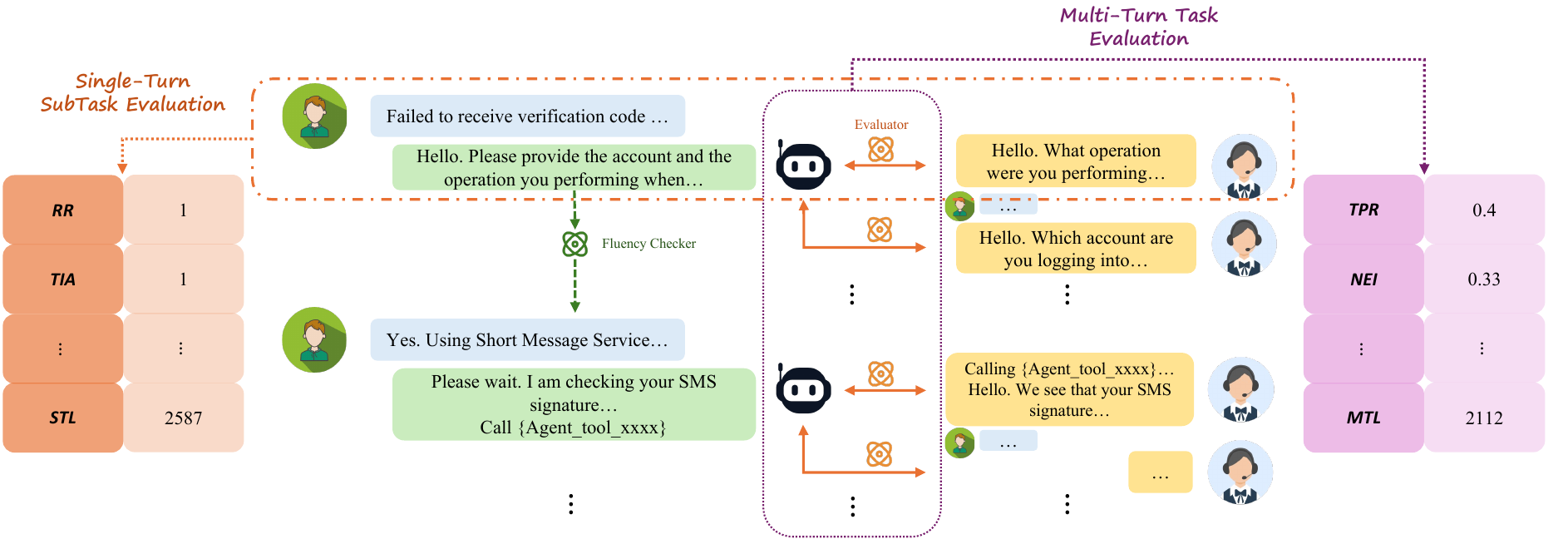}
      \caption{\textbf{The evaluation framework of CirrusBench.} The evaluation of a multi-turn task proceeds as a sequence of single-turn subtask sessions.}
      \label{fig:eval_Cirrus}
\end{figure*}

\subsubsection{Single-Turn Subtask Evaluation}

The foundation of our benchmark is the granular evaluation of an agent's response within a single conversational turn. This framework allows for a precise diagnosis of the agent's capabilities across three critical dimensions: Tool-Use Competency, Problem-Solving Ability, and Customer Service Qualification.

     \textbf{Tool-Use Competency}. When an agent decides to use a tool, its action is not treated as a monolithic success or failure. Instead, we evaluate it as a cascading sequence of decisions, each with its own verifiable outcome: (1) \textit{Tool Invocation Awareness (TIA)}: Describes the correctness of the agent's decision regarding the necessity of a tool call. (2) \textit{Tool Selection Accuracy (TSA)}: Evaluates whether the agent selected the correct tool when a tool was required. (3) \textit{Tool Execution Validity (TEV)}: Measures the agent's ability to correctly format the inputs for the selected tool, ensuring successful execution.

     \textbf{Problem-Solving Ability}. In many Cloud service scenarios, the optimal response is not a tool call but a direct, informative answer. For the reply component, we evaluate: \textit{Reply Resolvability (RR)}: Determines whether the agent's natural language response contains the key information, conclusions, or instructions necessary to resolve the user's query at that specific step. This ensures the agent is not just conversational but is actively driving towards a solution.

     \textbf{Customer-Centric Service Qualification Metric}. Distinct from functional correctness, we define the \textit{Single-Turn Latency (STL)} to quantify the quality and efficiency of the agent's service delivery, which are paramount in the cloud service domain. We define $STL$ as the count of output tokens generated for the specific subtask inspired by the studies of LLM latency \cite{miao2025towards, kwon2023efficient}. The importance of low latency is also illustrated with an example in Fig. \ref{fig:case_with_latency} in Appendix. \ref{sec:appendix_case}.

    \subsubsection{Dynamic Multi-Turn Task Evaluation}
    While single-turn subtask evaluation is crucial for diagnostics, real-world problems often require a sequence of interactions. To evaluate an agent's performance over a complete ticket, we introduce a dynamic assessment mechanism that leverages historical solution paths but allows for deviation.
    
    The multi-turn process, illustrated in Fig. \ref{fig:eval_Cirrus} and Algorithm \ref{alg:eval} in the Appendix. \ref{sec:appendix_multiturn}, operates as follows: At each beginning checkpoint of a single-turn subtask evaluation session, the agent receives the conversation history and the current customer's query to generate a response, which includes both a reply component and a tool call component. The single-turn subtask evaluation compares the agent’s response against the ground truth to compute single-turn metrics, such as Reply Resolvability (RR) and Tool Invocation Awareness (TIA), using an evaluator and a critic function. This single-turn subtask evaluation repeats for the ground truth of subsequent checkpoints until the evaluation returns a "fail to accept" result. Upon the "fail to accept" result, the current single-turn subtask evaluation terminates, then a Fluency Checker determines whether the subsequent customer query of the failed checkpoint is logically valid given the agent's deviant response. If the Fluency Checker returns True, a new single-turn subtask session is initiated with the current customer's query and updated history messages built upon the historical agent's responses and the customer's queries; otherwise, the multi-turn task evaluation terminates. When the multi-turn task evaluation terminated, global metrics such as Task Progression Rate (TPR), Normalized Efficiency Index (NEI), and Multi-Turn Latency (MTL) are computed.

    \textbf{Metrics for Multi-Turn task}. Based on the metrics obtained from single-turn subtask evaluation, we introduce metrics for multi-turn task evaluation to measure the problem-solving ability, efficiency, and latency of the agent. We define: (1) \textit{Task Progression Rate (TPR)}: Describes the proportion of successfully resolved single-turn subtasks within a multi-turn task. It is defined as the ratio of the number of successfully resolved single-turn subtasks to the total number of single-turn subtasks within the multi-turn task. (2) \textit{Logical Jump (LJ)}: Represents the number of skipped single-turn subtasks before the evaluation of the multi-turn task terminates. It takes a value in $\{0, \dots, k-1\}$, where $k$ is the total number of subtasks successfully processed until termination. (3) \textit{Normalized Efficiency Index (NEI)}: Evaluates the agent's efficiency in resolving the multi-turn task, defined as the ratio of the Logical Jump (LJ) to the total number of subtasks successfully processed ($k$), that is $NEI = \frac{LJ}{k-1}$ if $k > 1$; $NEI=1$ if $k=1$; otherwise $NEI=0$. (4) \textit{Multi-Turn Latency (MTL)}: Defined as the average of the Single-Turn Latency (STL) across all successfully processed single-turn subtasks within the multi-turn task.

\section{Experiments}
\label{sec:exp}
\subsection{Evaluation Settings}

We conduct experiments on CirrusBench , which comprises the Qwen series \cite{yang2025qwen3}, GPT series \cite{achiam2023gpt} and DeepSeek series \cite{liu2024deepseek}. All LLMs are evaluated via corresponding APIs.

\subsection{Evaluation Metrics}

The evaluation metrics are average of the multi-turn task evaluation metrics, including: \textit{(1) Average Task Progression Rate (ATPR)}, the average of TPR on the dataset. \textit{(2) Average Logical Jump (ALJ)}, the average of LJ on the dataset. \textit{(3) Average Normalized Efficiency Index (ANEI)}, the average of NEI on the dataset. \textit{(4) Average Multi-Turn Latency (AMTL)}, the average of MTL on the dataset. And we also include \textit{(5) 
Success Rate (SR)}, the rate of successfully resolved multi-turn tasks, which is evaluated on both pass@1 score and pass@2 score.

\subsection{ Reliability of Evaluator}
For the evaluator used in the single-turn subtask evaluation to judge whether the agent-generated response covers the key informations of ground truth response, we leverage DeepSeek-V3.2 and a carefully engineered prompt. As shown in Table. \ref{tab:confusion_matrix}, the evaluator achieves an accuracy of 91.49\% on a validation set of 141 annotations provided by human experts, demonstrating the reliability of the proposed judge. For stability of altering judge's base model, we altering the base model with GPT-5, DeepSeek-R1, GPT-5.2, Gemini-3-pro, the difference on accuracy of those base models are within 3\%. For the benefit of cost, we adopt DeepSeek-V3.2 as our evaluator's base model.

\subsection{Main Results}

\subsubsection{Results on CirrusBench}

\begin{figure}[h!] %
      \centering
      \includegraphics[width=0.9\linewidth]{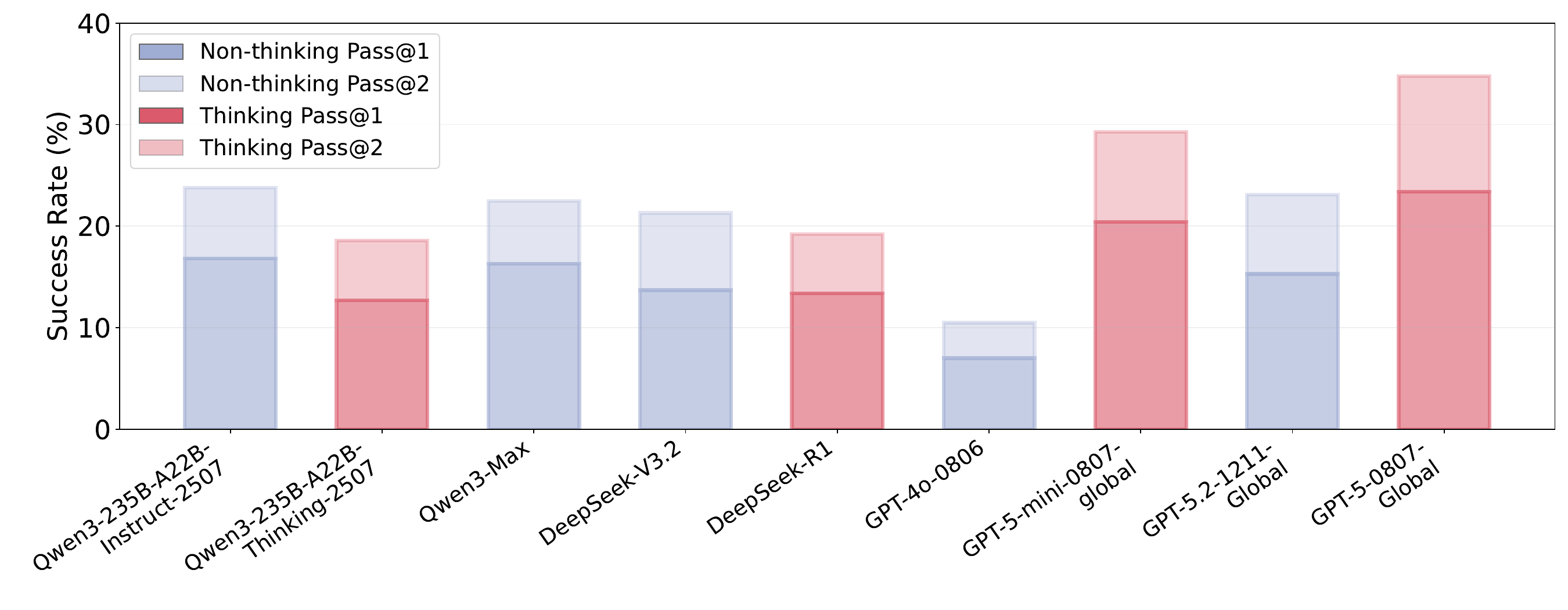}
      \caption{\textbf{Overall Success Rate results on CirrusBench.}}
      \label{fig:success_rate_on_CirrusBench}
\end{figure}

We compute the overall comprehensive evaluation results on CirrusBench in the following two different settings. Firstly, We evaluate different LLM models on the dataset with tool-call tasks and the dataset without tool-call tasks separately. As shown in Table. \ref{tab:model_performance}, in almost all metrics, the performances on the dataset with tool-call tasks are better than on the dataset without tool-call tasks, which demonstrates that the ability for tool-use is a challenge for agent. In GPT series, the models with explicit thinking capability perform better than the models without thinking capability in SR, ATPR, ALJ and ANEI, but cost more in the latency metric AMTL. However, explicit thinking capability is not always benefit for SR, ATPR, ALJ and ANEI, which can be observed from the Qwen series and DeepSeek series. For instance, the thinking model DeepSeek-R1 achieves a Pass@1 Success Rate (SR) of only 7.9\% in tool-use scenarios, which is lower than the 9.6\% achieved by its non-thinking counterpart, DeepSeek-V3.2. Similarly, Qwen2.5-72B-Thinking achieves an SR of 11.0\% in tool scenarios, showing no improvement over the standard Qwen2.5-72B-Instruct (11.0\%), while performing significantly worse in non-tool scenarios (13.3\% vs. 19.0\%). Moreover, the results underscore the trade-offs between model architectures that while standard models like GPT-4o-0806 maintain extremely low latency (188.6/ 250.8 tokens), they may struggle with the absolute success rates (Pass@1 SR 6.8\%/ 7.0 \%) compared to larger, more computationally intensive models. In Fig. \ref{fig:success_rate_on_CirrusBench}, the SR results are obtained by evaluating these LLM models on the whole dataset of CirrusBench, merging the tool-call tasks and non-tool-call tasks, which is consistent with the evaluation results in Table. \ref{tab:model_performance}.

To analyze the impact of different factors of CirrusBench, we derived experiments in dataset with different input tokens, different numbers of checkpoints, different service categories and different reply types. The results are illustrated in the following.

\begin{table*}[h!]
\caption{Evaluation of various models across dataset with tool-call requirement and dataset without tool-call requirement of CirrusBench. Red color indicates models with explicit thinking capabilities. Bold values indicate the best performance in each column, and underlined values indicate the worst performance in each column.}
\centering
\resizebox{\textwidth}{!}{
\begin{tabular}{lcccccccccccccccc}
\toprule
\multirow{2}{*}{\textbf{Models}} & \multicolumn{2}{c}{\textbf{Pass@1 SR (\%)} $\uparrow$} & \multicolumn{2}{c}{\textbf{Pass@2 SR (\%)} $\uparrow$} & \multicolumn{2}{c}{\textbf{ATPR (\%)} $\uparrow$} & \multicolumn{2}{c}{\textbf{ALJ} $\uparrow$} & \multicolumn{2}{c}{\textbf{ANEI (\%)}  $\uparrow$} & \multicolumn{2}{c}{\textbf{Avg. All Tokens}} & \multicolumn{2}{c}{\textbf{Avg. Input Tokens}} & \multicolumn{2}{c}{\textbf{AMTL (tokens)} $\downarrow$} \\
\cmidrule(lr){2-3} \cmidrule(lr){4-5} \cmidrule(lr){6-7} \cmidrule(lr){8-9} \cmidrule(lr){10-11} \cmidrule(lr){12-13} \cmidrule(lr){14-15} \cmidrule(lr){16-17}
 &  w tool & w/o tool & w tool & w/o tool & w tool & w/o tool & w tool & w/o tool & w tool & w/o tool & w tool & w/o tool & w tool & w/o tool & w tool & w/o tool \\
\midrule
Qwen3-235B-A22B-Instruct-2507 & 11.0 & 19.0 & 17.4 & 26.2 & 25.6 & 34.2 & 0.101 & 0.156 & 12.3 & 31.5 & 11900.1 & 11714.3 & 11710.0 & 11120.6 & 190.1 & 593.7 \\
{\color{red} Qwen3-235B-A22B-Thinking-2507} & 11.0 & 13.3 & 18.8 & 18.5 & 27.8 & 24.8 & 0.106 & 0.090 & 12.5 & 23.7 & 13659.2 & 12013.9 & 12596.1 & 10973.3 & 1063.1 & 1040.5 \\
Qwen3-Max & 14.3 & 17.1 & 19.4 & 23.7 & 28.6 & 32.2 & 0.108 & 0.137 & 12.5 & 31.8 & 12643.3 & 11666.3 & 12409.7 & 11194.5 & 233.6 & 471.8 \\
DeepSeek-V3.2 & 9.6 & 15.2 & 15.5 & 23.5 & 21.9 & 31.4 & 0.104 & 0.138 & 12.5 & 31.5 & 12006.2 & 11577.0 & 11813.4 & 11060.5 & 192.8 & 516.5 \\
{\color{red} DeepSeek-R1} & 7.9 & 15.2 & 10.1 & 21.1 & 21.1 & 29.3 & 0.082 & 0.118 & 11.1 & 29.0 & 16039.5 & 11748.2 & 14565.3 & 10881.1 & 1474.2 & 867.2 \\
GPT-4o-0806 & \underline{6.8} & \underline{7.0} & \underline{8.7} & \underline{11.0} & \underline{15.0} & \underline{16.0} & \underline{0.042} & \underline{0.045} & 10.9 & \underline{18.6} & 12156.2 & 11318.8 & 11978.6 & 11067.9 & \textbf{177.6} & \textbf{250.8} \\
{\color{red} GPT-5-mini-0807-global} & 15.2 & 22.1 & 22.6 & 31.4 & 35.4 & 40.9 & 0.167 & 0.193 & \textbf{20.0} & 37.7 & 13553.1 & 13337.6 & 12323.8 & 11887.2 & 1229.3 & 1450.4 \\
GPT-5.2-1211-Global & 10.3 & 16.8 & 16.3 & 25.2 & 27.3 & 34.3 & 0.116 & 0.125 & 12.7 & 32.1 & 12657.5 & 11722.1 & 12388.9 & 11349.3 & 268.6 & 372.8 \\
{\color{red} GPT-5-0807-Global} & \textbf{16.7} & \textbf{23.8} & \textbf{27.8} & \textbf{35.2} & \textbf{39.8} & \textbf{44.0} &\textbf{0.261} & \textbf{0.258} & \underline{10.7} & \textbf{37.8} & 25433.0 & 14282.0 & 23850.0 & 12047.6 & \underline{1583.0} & \underline{2234.4} \\
\bottomrule
\end{tabular}
}

\label{tab:model_performance}
\end{table*}

\begin{table}[h!]
\centering
\caption{Model performance across input token ranges. (on GPT-5)}
\label{tab:acc_tokens}
\resizebox{0.48\textwidth}{!}{%
\begin{tabular}{lccccc}
\toprule
\textbf{Token Range}  & \textbf{pass@1 SR (\%) $\uparrow$} & \textbf{pass@2 SR (\%) $\uparrow$} &\textbf{TIA } $\uparrow$&\textbf{TSA} $\uparrow$&\textbf{TEV} $\uparrow$\\ \midrule
0 -- 8,000                    & 40.4        & 60.7   & 0.531 & 0.477 & 0.261\\
8,000 -- 16,000               & 43.2        & 63.4   & 0.535 & 0.441 & 0.244\\
Over 16,000                   & 44.8        & 64.8   & 0.536 & 0.451 & 0.267\\
\bottomrule
\end{tabular}
}
\end{table}

\begin{table}[h!]
\caption{Model performance across different counts of checkpoints. (on GPT-5)}
\label{tab:tpr_checkpoints}
\centering
\resizebox{0.48\textwidth}{!}{%
\begin{tabular}{lcccccc}
\toprule
\textbf{Count of Checkpoints}  & \textbf{Pass@1 SR(\%) $\uparrow$}& \textbf{Pass@2 SR(\%) $\uparrow$} & \textbf{TPR (\%) $\uparrow$} & \textbf{ALJ $\uparrow$ } &\textbf{ANEI} $\uparrow$ & \textbf{AMTL (tokens)} $\downarrow$\\ 
\midrule
1         &   45.8   & 58.5  & 58.5  & 0     &  0.58 & 2410\\
2          &    28.1 & 39.6  &47.1   &  0.19 & 0.33  & 2270\\
3           &   17.4 & 29.5  & 42.0  & 0.28  &  0.30 & 2237\\
4            &  9.2 & 19.2  & 32.9  & 0.40  &  0.31 & 2190\\
Others (5+ )  & 5.3 & 14.6  & 15.7  & 0.59  &0.35& 2151\\ 
\bottomrule
\end{tabular}
}

\end{table}

\begin{table}[h!]
\caption{Model performance across service categories. (on GPT-5)}
\label{tab:tpr_product}
\centering
\resizebox{0.48\textwidth}{!}{%
\begin{tabular}{lcccccc}
\toprule
\textbf{Category} & \textbf{Pass@1 SR(\%) $\uparrow$}& \textbf{Pass@2 SR(\%) $\uparrow$} & \textbf{TPR (\%) $\uparrow$} & \textbf{ALJ $\uparrow$ } &\textbf{ANEI} $\uparrow$ & \textbf{AMTL (tokens)} $\downarrow$\\ 
\midrule
Server       & 21.6  & 31.4  & 41.0  & 0.319 &0.364&2304\\
Registration  & 19.4 & 32.0  & 40.6  & 0.286 &0.322&2099\\
SMS Services & 20.6  & 34.0  & 41.8  & 0.248 &0.358&2393\\
Website      &  36.8 & 49.1  & 54.9  & 0.281 &0.465&2218\\
Email        & 18.9  & 27.5  & 36.7  & 0.172 &0.347&2190 \\
Others (15+ )& 26.7  & 37.8  & 47.2  & 0.230 &0.409&2228\\ 

\bottomrule
\end{tabular}
}
\end{table}

\begin{figure}[h!] %
       \includegraphics[width=0.8\linewidth]{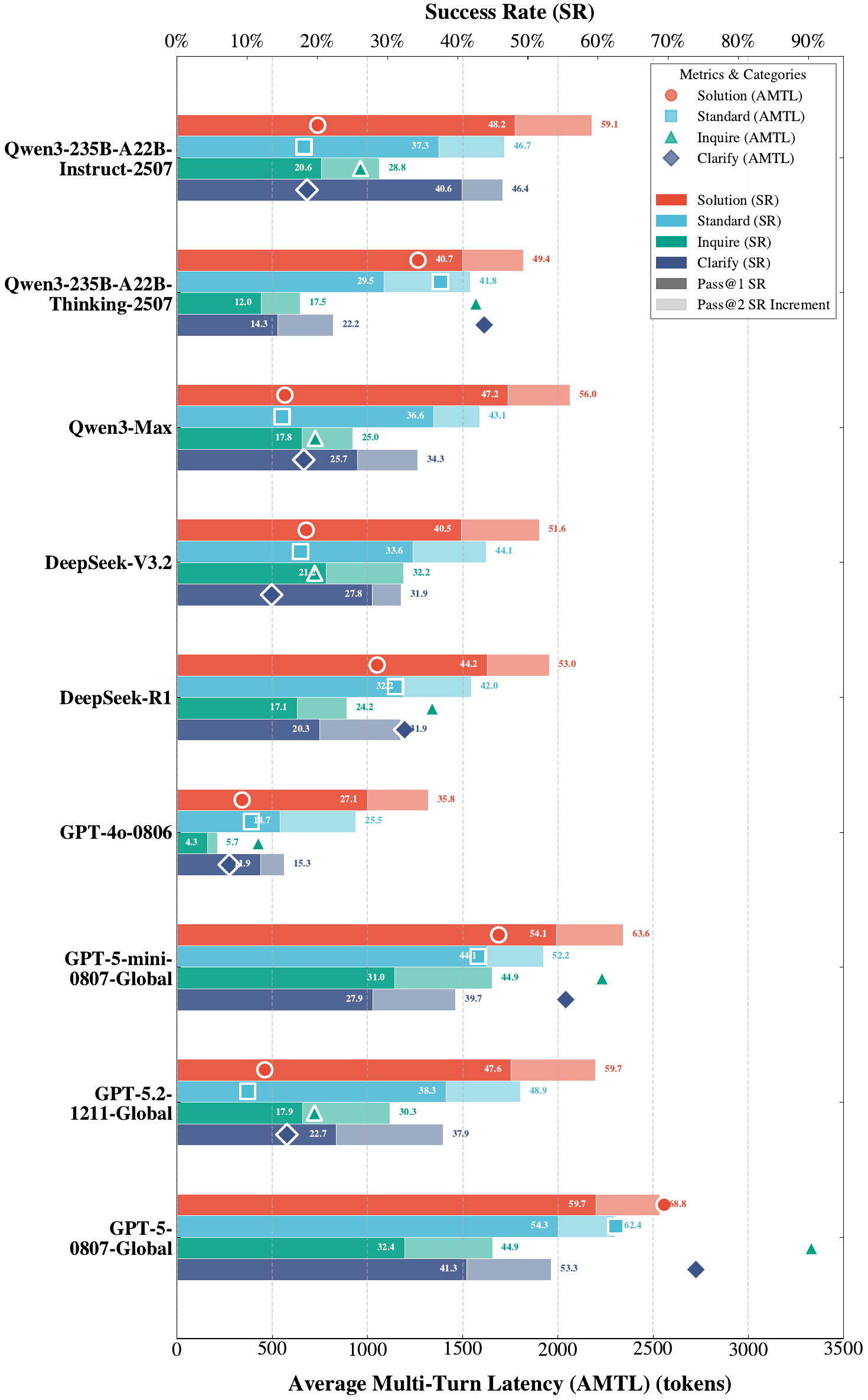}
        \caption{\textbf{Pass@1, pass@2 success rate and AMTL by different reply types}.}
        \label{fig:reply_type}
\end{figure}

\subsubsection{Impact of Input Token Count}

We evaluate the impact of input token count through single-turn subtask evalution methodology. Table \ref{tab:acc_tokens} illustrates the model's performance across varying input context lengths. Contrary to the "lost-in-the-middle" phenomenon often observed in general NLP tasks, our results on CirrusBench reveal a positive correlation between context length and resolution success. The model achieves its peak performance in the \textbf{Over 16,000} token range, with a Pass@1 of 44.8\% and a Pass@2 of 64.8\%, significantly outperforming the 0--8,000 token baseline (40.4\%).

This trend supports the hypothesis that domain-specific professional knowledge and sufficient user information are necessary conditions for LLMs to correctly resolve complex service tasks. In specialized customer service, issues are inherently intricate, often depending on specific system logs, version compatibility, and detailed error tracebacks. Short inputs may lack these critical constraints, leading to hallucinated or generic advice. Conversely, extended contexts provide the comprehensive information density required to ground the model's reasoning, allowing it to leverage its professional training data effectively. Thus, the ability to process long-context inputs is not merely an efficiency metric but a prerequisite for accuracy in this domain.

\subsubsection{Impact of Checkpoints Count}

To investigate the robustness of LLMs over extended interactions, we categorize model performance based on the count of checkpoints required to reach a final resolution. The results, summarized in Table \ref{tab:tpr_checkpoints}, reveal a clear inverse correlation between interaction depth and task success.

\begin{itemize} 
\item \textbf{Complexity of Sustained Reasoning:} Tasks with more checkpoints show lower SR and lower TPR. This suggests that while model is capable of "one-shot" problem solving, they struggle to maintain logical consistency and state tracking across extended, multi-turn diagnostic sessions. 
\item \textbf{Risky Exploration:} As the interaction depth grows, the ALJ increases but the ANEI drops. This suggests that while model try to explore paths with deviation but obtains much more risk in this exploration, which leads to wrong short path more than reliable resolution path. 
\item \textbf{Error Accumulation:} Minor reasoning errors in early turns propagate and compound, causing the model to deviate irreversibly from the correct solution path over time. 
\end{itemize}

In summary,  multi-turn consistency remains a critical bottleneck. The ability to maintain logical grounding and resist error accumulation across extended diagnostic sessions is a prerequisite for deploying autonomous agents in professional service environments.

\subsubsection{Impact of Service categories}

Table \ref{tab:tpr_product} presents the performance of the model across various Cloud service categories. The results reveal a significant performance gap between categories (for instance, pass@1 SR is 36.8\% in Website category while 18.9\% in Email category), reflecting the inherent technical complexity and heterogeneous nature of real-world cloud troubleshooting scenarios.

The performance variance across categories underscores that a "one-size-fits-all" approach is insufficient for service agents,which validates the necessity of CirrusBench's multi-category design, as it accurately captures the diverse challenges that autonomous agents must overcome in professional service environments.

\subsubsection{Performance Analysis by Response Type}

As shown in Figure. \ref{fig:reply_type}, a comparative analysis of the four response types reveals a substantial performance disparity. The evaluated models exhibit their highest proficiency in producing Solution-type responses, whereas their performance on the Inquiry response type is consistently the weakest.

This pattern indicates that, when operating as service agents, LLMs display a systematic tendency toward premature closure. Instead of engaging in a structured Clarify phase to resolve ambiguities, the models frequently bypass critical information-gathering steps and proceed directly to proposing solutions despite having insufficient contextual information. We attribute this behavior to two principal capability limitations in current LLMs:

1. Deficiency in Uncertainty Identification: The models exhibit difficulty in recognizing “information lacunae” (i.e., missing or incomplete information) and often fail to detect when a user’s intent is too underspecified to support reliable action.

2. Lack of Proactive Strategy: The models lack robust, goal-directed, and proactive questioning strategies that would systematically guide the dialogue toward a successful, closed-loop service interaction.

Additionally, Figure \ref{fig:outputtokentypes} demonstrates that producing Inquiry and Clarify responses generally incurs a higher average token cost. Given their comparatively low success rates, this increased computational expenditure indirectly reflects the models’ difficulty in handling these response types. The elevated token count suggests that generating precise, diagnostic questions imposes a greater cognitive load on the model’s reasoning processes than does synthesizing a direct solution. This further supports the conclusion that interactive information seeking remains a critical bottleneck for the development of effective service-oriented LLM-based agents.






\section{Conclusion}
\label{sec:conclusion}
In this work, we introduced CirrusBench, the first evaluation framework systematically derived from authentic, real-world cloud service tickets. Unlike previous benchmarks that rely on synthetic environments or role-playing, CirrusBench preserves the high entropy, long-horizon dependencies, and complex tool usage inherent in actual technical service interactions. By shifting the evaluation paradigm from simple execution correctness to Customer-Centric Metrics, including metrics such as TPR, NEI, and MTL, we provide a more rigorous standard for assessing an agent's practical viability.

Our experimental analysis highlight a substantial gap between the current capabilities of state-of-the-art LLMs and the rigorous demands of real-world cloud service support. While models exhibit strong reasoning potential, they face a critical bottleneck in tool integration, where performance suffers a sharp decline compared to pure conversational tasks. The study further reveals a complex trade-off regarding "thinking" models; while explicit reasoning capabilities can be beneficial, they frequently introduce prohibitive latency penalties without yielding proportional improvements in problem solving, effectively compromising the efficiency required for customer satisfaction. And the knowledge embedding shows importance for agent in Cloud service tasks. Additionally, models demonstrated significant fragility in long-horizon interactions, with problem solving metrics dropping as the number of checkpoints increased due to error accumulation and a failure to maintain logical consistency. This is compounded by a tendency toward premature closure, as agents consistently underperformed in "Inquiry" tasks, failing to proactively ask necessary clarifying questions before attempting a solution. Ultimately, these findings indicate that future development must move beyond isolated accuracy metrics to prioritize resolution efficiency, multi-turn robustness, and responsive information-gathering strategies.

Ultimately, CirrusBench underscores that for LLM agents to succeed in cloud service support, future research must move beyond optimizing for accuracy in isolation. It is imperative to develop agents that are not only factually correct and proficient in tool use but also responsive and efficient enough to maintain customer trust before patience is exhausted. We hope this benchmark serves as a catalyst for the development of next-generation agents that truly align with the dynamic needs of human customers.

\section{Acknowledgments}
\label{sec:acknowledgments}
To Zhiqiang Liu, Aiyu Chen, Ning Zhang of Alibaba Group, for translating the original Cloud Service tickets into English.

\section*{Impact Statement}

This paper introduces a benchmark aimed at advancing research on Large Language Models (LLMs) agent in the application of Cloud Service environment. It is important to acknowledge that our experiments and evaluations rely heavily on LLMs, but this study does not fully explore or mitigate potential biases inherent in their outputs. Addressing these biases and ensuring model alignment with social values remain critical challenges. This underscores the importance of conducting comprehensive evaluations that consider diverse dimensions of human society and their implications, none of which we feel must be specifically highlighted here.

\bibliographystyle{unsrt}  
\bibliography{main.bib}  


\newpage
\appendix
\label{sec:appendix}
\onecolumn
\section{Definition of Metrics}\label{sec:appendix_metrics}

\subsection{Single-Turn Evaluation Metrics}
For single-turn evaluation, we define the following metrics:

\begin{description}
    \item[Tool Invocation Awareness ($TIA$):] Describes the correctness of the agent's decision regarding the necessity of a tool call. We define $TIA = 1$ if the agent correctly invokes a tool when required, or correctly abstains from calling a tool when none is needed; otherwise, $TIA = 0$.

    \item[Tool Selection Accuracy ($TSA$):] Evaluates whether the agent selected the correct tool from the available toolset when a tool was required. This tests the agent's ability to differentiate between tool functionalities. If the tool selection is correct, $TSA = 1$; otherwise, $TSA = 0$.
    
    \item[Tool Execution Validity ($TEV$):] Measures the agent's ability to correctly format the inputs for the selected tool, ensuring successful execution. If the tool invocation is successful, $TEV = 1$; otherwise, $TEV = 0$.
    
    \item[Reply Resolvability ($RR$):] Determines whether the agent's natural language response contains the key information, conclusions, or instructions necessary to resolve the user's query at that specific step.
    
    \item[Single-Turn Latency ($STL$) $\downarrow$:] Quantifies the latency based on the computational cost of the response. We define $STL$ as the count of output tokens generated for the specific subtask.
\end{description}

\subsection{Multi-Turn Evaluation Metrics}
A multi-turn task consists of a sequence of single-turn subtasks arranged chronologically. For multi-turn evaluation, we define the following metrics:

\begin{description}
    \item[Task Progression Rate ($TPR$) $\uparrow$:] Describes the proportion of successfully resolved single-turn subtasks within a multi-turn task. It is defined as:
    \[
    TPR = \frac{\text{number of successfully resolved single-turn subtasks}}{\text{total number of single-turn subtasks in the multi-turn task}}
    \]
    
    \item[Logical Jump ($LJ$) $\uparrow$:] Represents the number of skipped single-turn subtasks before the evaluation of the multi-turn task terminates. It takes a value in $\{0, \dots, k-1\}$, where $k$ is the total number of subtasks successfully processed until termination.
    
    \item[Normalized Efficiency Index ($NEI$) $\uparrow$:] Evaluates the agent's efficiency in resolving the multi-turn task, defined as the ratio of the Logical Jump ($LJ$) to the total number of subtasks successfully processed ($k$). Specifically, if the evaluation terminates after successfully processed $k$ subtasks:
    \[
    NEI = 
    \begin{cases} 
    \frac{LJ}{k - 1} & \text{if } k > 1 \\
    1 & \text{if } k = 1 \\
    0 & \text{otherwise}
    \end{cases}
    \]
    
    \item[Multi-Turn Latency ($MTL$) $\downarrow$:] Defined as the average of the Single-Turn Latency ($STL$) across all successfully processed single-turn subtasks within the multi-turn task.
\end{description}

\subsection{Dataset Evaluation Metrics}
A dataset consists of multiple multi-turn tasks. For the evaluation of the dataset, we define the following metrics:

\begin{description}
    \item[Average Task Progression Rate ($ATPR$) $\uparrow$:] The average $TPR$ across all multi-turn tasks in the dataset, taking a value in $[0, 1]$.
    
    \item[Average Logical Jump ($ALJ$) $\uparrow$:] The average $LJ$ across all multi-turn tasks where at least one subtask was successfully resolved. Formally:
    \[
    ALJ = \frac{\sum_{\text{tasks with successful subtasks}} LJ}{\text{number of tasks with successful subtasks}}
    \]
    
    \item[Average Normalized Efficiency Index ($ANEI$) $\uparrow$:] The average $NEI$ across all multi-turn tasks where at least one subtask was successfully resolved. Formally:
    \[
    ANEI = \frac{\sum_{\text{tasks with successful subtasks}} NEI}{\text{number of tasks with successful subtasks}}
    \]
    
    \item[Success Rate ($SR$) $\uparrow$:] Describes the proportion of multi-turn tasks that are completely resolved (i.e., tasks where $TPR = 1$). We define:
    \[
    SR = \frac{\text{count of completely resolved tasks}}{\text{total count of tasks in the dataset}}
    \]
    
    \item[Average Multi-Turn Latency ($AMTL$) $\downarrow$:] Describes the average Multi-Turn Latency across all multi-turn tasks in the dataset where at least one subtask was successfully resolved. We define:
    \[
    AMTL = \frac{\sum_{\text{tasks with successful subtasks}} MTL}{\text{number of tasks with successful subtasks}}
    \]
\end{description}

\section{The Multi-Turn Task Evaluation Procedure of CirrusBench}\label{sec:appendix_multiturn}
The multi-turn task evaluation procedure of CirrusBench is decribed in Section \ref{sec:methodology}. Here we demonstrate the detail of it in Algorithm. \ref{alg:eval}.

\begin{algorithm}[tb]
    \caption{CirrusBench evaluation for one Multi-Turn task}
    \label{alg:eval}
    \begin{algorithmic}
        \State {\bfseries Input:}
        {$Agent$ \# Agent to be evaluated \\
        $Evaluator$ \# Judging whether the reply of agent covers the key informations of the ground truth reply \\
        $FluencyChecker$ \# Checking whether the next query is available in logical fluency \\
        $D = \left[H, (U_1, R_1, T_1), \dots, (U_K, R_K, T_K)\right]$ \# $K$ key checkpoints of dialog data}
        \State Initialize $h \gets H$  \# Conversation history
        \State Initialize $pass \gets True$ \# Pass flag
        \State Initialize $SucResp \gets 0$ \# Counts of successful response
        \State Initialize $MTL \gets 0$ \# Multi-Turn Latency
        \While{$1\le k \le K$ and $fluency$ is $True$} 
            \State $(\hat{R}_k, \hat{T}_k), STL \gets Agent(H, U_k)$ 
            \State $MTL \gets MTL + STL$
            \State $i \gets k - 1$
            \While{$pass$ is $True$ and $i < K$}
                \State $i \gets i+1$
                \If{$i < K$}
                    \State $pass \gets Evaluator\left((\hat{R}_k, \hat{T}_k), (R_i, T_i)\right)$
                \Else
                    \State $pass \gets False$
                \EndIf
            \EndWhile
            \If{$i > k$}
                \State $SucResp \gets SucResp + 1$
                \State $h$.extend$\left((U_k, \hat{R}_k, \hat{T}_k)\right)$
                \State $k \gets i$
                \State $fluency \gets FluencyChecker(H, U_k)$
            \Else
                \State $fluency \gets False$
            \EndIf
        \EndWhile
        \State $k \gets k-1$
        \State $LJ \gets k-SucResp$
        \If{$k=1$}
            \State $NEI \gets 1$
        \ElsIf{$k=0$}
            \State $NEI \gets 0$
        \Else
            \State $NEI \gets \frac{LJ}{k-1}$
        \EndIf   
        \State $TPR \gets \frac{k}{K}$ \\
        \State $MTL \gets \frac{MTL}{k}$ \\
        \State {\bfseries Output:} 
            $LJ, NEI, TPR, MTL$    
    \end{algorithmic}
\end{algorithm}

\section{The Reliability of Evaluator}\label{sec:appendix_evaluator}
In this evaluator, we compare whether the model-generated response covers the key informations of human-authored response. Using DeepSeek-V3.2 and a carefully engineered prompt, as shown in Table. \ref{tab:confusion_matrix}, the evaluator achieves an accuracy of 91.49\% on a validation set of 141 annotations provided by human experts, demonstrating the reliability of the proposed judge. For stability of altering judge's base model, we altering the base model with GPT-5, DeepSeek-R1, GPT-5.2, Gemini-3-pro, the difference on accuracy of those base models are within 3\%. For the benefit of cost, we adopt DeepSeek-V3.2 as our evaluator's base model. 

\begin{table}[t!]
    \centering
    \caption{Confusion Matrix Results of evaluator}
    \label{tab:confusion_matrix}
    \begin{tabular}{lcc} 
        \toprule
        \textbf{Actual / Predicted} & \textbf{Positive} & \textbf{Negative (\%)} \\ 
        \midrule
        \textbf{Positive}   & 60 & 9 \\
        \textbf{Negative}   & 3 & 69 \\
        \bottomrule
    \end{tabular}
\end{table}

\section{Examples of CirrusBench Data}\label{sec:appendix_case}

\begin{figure}[t] %
    \begin{minipage}[t]{0.45\textwidth}
        \includegraphics[width=\linewidth]{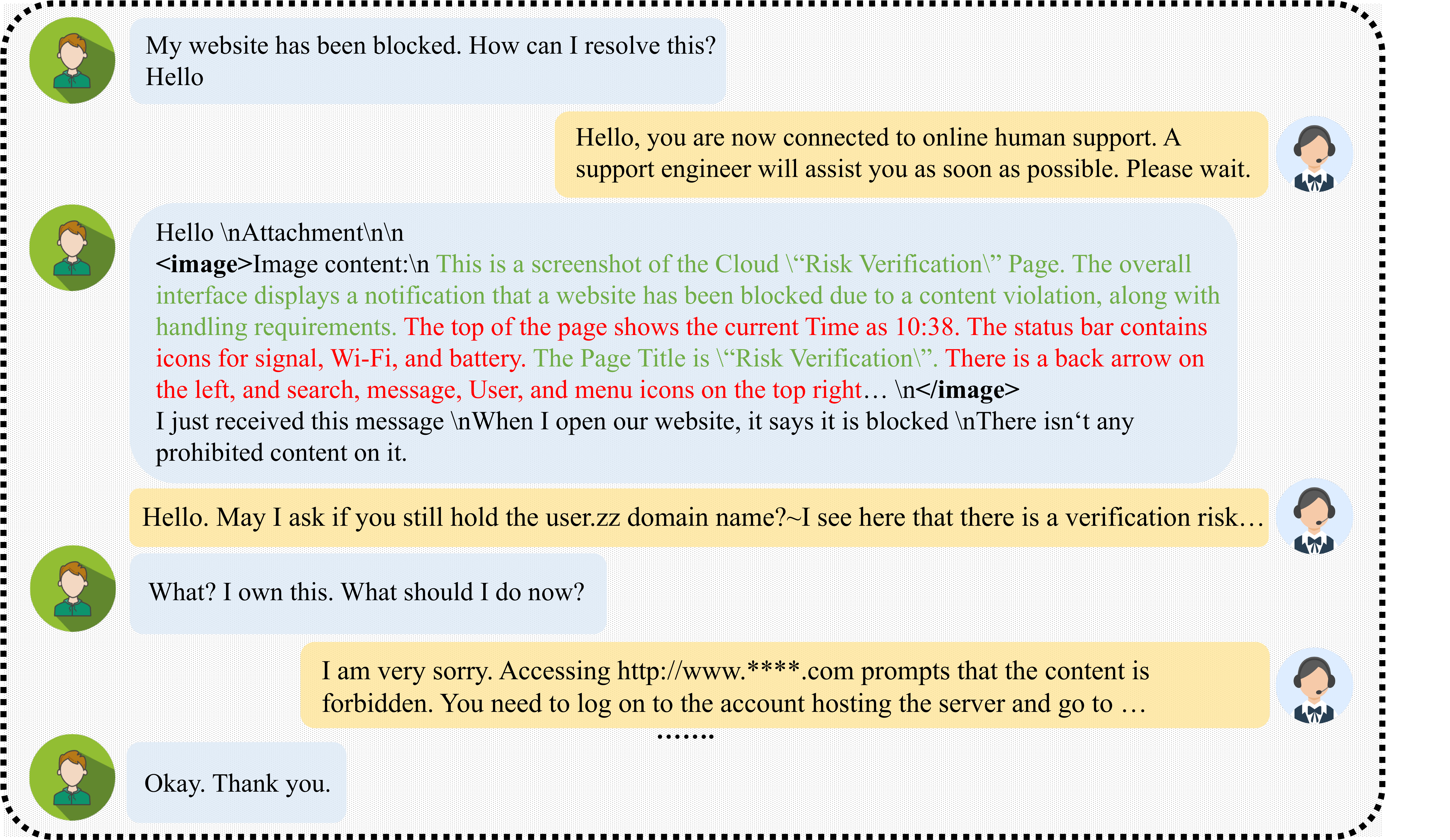}
        \caption{\textbf{Case Study of Customer's High-Entropy Input.} A real-world example where a customer supplements a blocked-website report with a raw screenshot. The image contains significant stochastic noise, such as signal strength and timestamps, orthogonal to the issue, requiring the agent to filter visual noise and extract relevant error context.}
        \label{fig:case_with_image}
    \end{minipage}\qquad \qquad
    \begin{minipage}[t]{0.45\textwidth}
        \includegraphics[width=\linewidth]{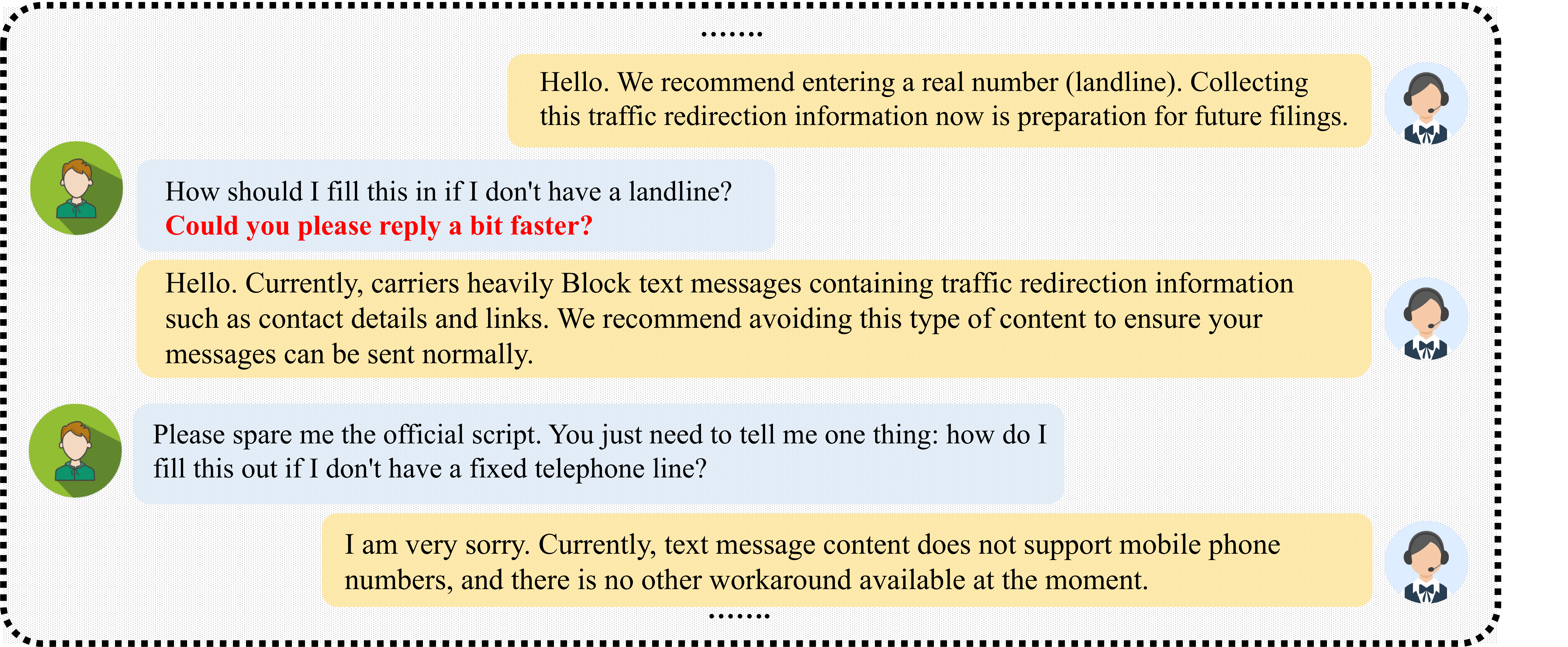}
        \caption{\textbf{Impact of Latency on User Sentiment.} A transcript illustrating the dependency between inference latency and customer agitation. The interaction demonstrates how delayed responses and rigid scripting cause a functional inquiry to devolve into an expression of dissatisfaction, validating the need for latency and sentiment-aware evaluation metrics.}
        \label{fig:case_with_latency}
    \end{minipage}
\end{figure}
Figure \ref{fig:case_with_image} exemplifies the high entropy inherent in real-world customer queries, which frequently necessitate robust multimodal processing capabilities. In this instance, the customer reports a service disruption regarding a blocked website, supplementing their textual query with a raw screenshot. Unlike synthetic datasets where visual inputs are often curated for relevance, this authentic input contains significant stochastic noise; the screenshot captures extraneous UI elements—such as battery status, signal strength, and system timestamps—that are semantically orthogonal to the reported content violation. This example demonstrates that robust service-oriented LLMs must possess the discriminative capacity to filter out such "visual noise" and extract the specific error context from complex, uncurated user uploads.

Figure \ref{fig:case_with_latency} underscores the critical dependency between inference latency and customer sentiment evolution in live support scenarios. The transcript reveals a temporal disconnect where the customer, frustrated by the interaction's pacing, explicitly demands a faster response. This latency, compounded by the agent’s reliance on rigid, official scripts, precipitates a visible escalation in customer agitation. Consequently, the interaction shifts from a functional inquiry regarding data entry to an expression of impatience and dissatisfaction. This trajectory validates the necessity for benchmarks to incorporate latency penalties and sentiment metrics; the ecological validity of a model is defined not merely by the correctness of its answer, but by its ability to deliver time-sensitive resolutions before user patience is exhausted.

\section{Prompts}
    \begin{tcolorbox}[
    enhanced, 
    colback=gray!5!white,
    colframe=gray!75!black,
    coltitle=white,
    fonttitle=\bfseries,
    title=Prompt used for response generate,
    arc=6pt, 
    boxrule=1pt,
    breakable, 
    width=\textwidth,
    sharp corners=south,
    drop shadow
]

\textbf{\texttt{<role\_mission>}}  

You are a professional customer support agent specializing in cloud computing products and services. Your primary mission is to resolve technical inquiries accurately, efficiently, and with high service quality while maintaining user trust.  

\textbf{\texttt{</role\_mission>}}

\textbf{\texttt{<task\_context>}}  

You are engaged in a support conversation where users submit queries regarding cloud service usage, configuration, billing, or troubleshooting. The system may provide contextual information such as user profile data, conversation history, or relevant documentation to aid problem resolution. You are expected to critically assess this context—determining its relevance—and leverage it judiciously in formulating your response.  

\textbf{\texttt{</task\_context>}}

\textbf{\texttt{<behavioral\_guidelines>}}  

\textbf{\texttt{<guideline>}}

Role fidelity: Consistently embody a human support agent. Never disclose AI identity, suggest external escalation, or reference internal system mechanics.

\textbf{\texttt{</guideline>}}  

\textbf{\texttt{<guideline>}}

Evidence grounding: All statements must derive from provided context, verified tools, or documented knowledge. Never fabricate facts, links, or parameters.

\textbf{\texttt{</guideline>}}  

\textbf{\texttt{<guideline>}}

Communication clarity: Lead with key conclusions. Structure multi-part responses with line breaks. Use natural, concise language without AI-typical markers (emojis, excessive formatting).

\textbf{\texttt{</guideline>}}  

\textbf{\texttt{<guideline>}}

User-centric interaction: Clarify ambiguities proactively. Prefer tool-assisted diagnosis over repeated user queries. Respond empathetically to sentiment without over-promising.

\textbf{\texttt{</guideline>}}  

\textbf{\texttt{</behavioral\_guidelines>}}

\textbf{\texttt{<tool\_usage>}}  

When domain-specific diagnostic tools are available and necessary: invoke only with fully satisfied parameters (user-provided or context-inferred); integrate results naturally into responses without naming tools or exposing invocation details.  

\textbf{\texttt{</tool\_usage>}}

\textbf{\texttt{<problem\_solving\_workflow>}}  

1. Analyze query intent and completeness.   
2. Audit available context for gaps.   
3. Select optimal action: resolve directly (if evidence sufficient), request minimal missing details (if tools cannot retrieve), invoke relevant tools (if parameters validated), or acknowledge scope limitations professionally.   
4. Deliver response aligned with behavioral guidelines.  

\textbf{\texttt{</problem\_solving\_workflow>}}

\textbf{\texttt{<information\_retrieval>}}  

Prioritize autonomous reasoning using provided references and tools before soliciting user input. Minimize unnecessary clarification cycles.  

\textbf{\texttt{<help\_doc>}}  
...  
\textbf{\texttt{</help\_doc>}}  

\textbf{\texttt{</information\_retrieval>}}

\end{tcolorbox}
    \begin{tcolorbox}[
    enhanced, 
    colback=gray!5!white,
    colframe=gray!75!black,
    coltitle=white,
    fonttitle=\bfseries,
    title=Prompt used for evaluator,
    arc=6pt, 
    boxrule=1pt,
    breakable, 
    width=\textwidth,
    sharp corners=south,
    drop shadow
]

\# Role  \\
You are a Senior Information Audit Expert, specialized in accurately analyzing the logical inclusion relationships between technical responses. Your task is to determine whether [Response B] substantively contains the core solution found in [Response A] regarding the user's problem.  \\
\\
\# Input Description  \\
History Info $<$recent\_messages$>$: Contains multiple rounds of dialogue between the customer and support; focus on the core needs of the user's final consecutive questions.  \\
Response A $<$content\_a$>$: A response to the final user question in the history.  \\
Response B $<$content\_b$>$: A response to the final user question in the history.  \\
\\
\# Output Format: Only output the final result; no other detailed metrics are required.\\
$<$result$>$\\
Included / Not Included\\
$<$/result$>$\\
\\
\# Task Description  \\
Determine whether [Response B] includes the content of [Response A]. \\ 
- **Included**: Response B covers the core technical facts, operational paths, or key conclusions in Response A that **solve the user's problem**.  \\
- **Not Included**: Response B is missing key operational instructions or core judgmental conclusions from Response A, or it provides a completely different technical path.  \\
\\
\# Audit Logic (Mandatory Chain of Thought)  \\
Before providing the result, please think through the following steps:  \\
1. **Intent Alignment**: Analyze the core needs of the user's final question.  \\
2. **Response A Core Extraction**: Identify the "core increment" in Response A that addresses the needs (e.g., implied premises like operational permission ownership, execution subjects, etc.).  \\
3. **Response B Comparison**: Check if Response B meets any of the following conditions:  \\
   - **Security Verification Coverage**:  \\
     - When Response A contains official identity verification markers (e.g., dedicated line numbers, ticket numbers), high-risk operation confirmation codes, or anti-fraud tips, Response B must retain this information in full.  \\
     - Exception: If Response B covers this via an equivalent official identifier (e.g., "Incoming call from xxx Cloud Filing Dedicated Line"), it is judged as Included.  \\
   - **Security Identification Coverage Principle**:  \\
     - For unique identifiers involving funds, account security, or official identity verification (e.g., outbound call segments/ticket numbers), missing these results in a judgment of "Not Included."

\end{tcolorbox}
    \begin{tcolorbox}[
    enhanced, 
    colback=gray!5!white,
    colframe=gray!75!black,
    coltitle=white,
    fonttitle=\bfseries,
    title=Prompt used for fluency checker,
    arc=6pt, 
    boxrule=1pt,
    breakable, 
    width=\textwidth,
    sharp corners=south,
    drop shadow
]

\# Role\\
You will play the role of a customer currently in a conversation with customer service. Based on the information provided below, determine whether the [Specified Question] can be asked following the [Historical Dialogue].\\

\# Input Description\\
The history content [Historical Dialogue]: Located within the $<$history$>$ tags.\\
The [Specified Question]: located within the $<$query$>$ tags.\\

\# Judgment Criteria:\\
If asking the [Specified Question] does not conflict with the [Historical Dialogue] in terms of logical validity and business continuity, determine: "Yes"; otherwise, determine: "No".\\

\# Output Format: \\
Please output only "Yes" or "No", and do not include any other text, punctuation, or explanation.

\end{tcolorbox}

\end{document}